\documentclass{article}
\usepackage{spconf,amsmath,graphicx}
\usepackage{algorithmicx,algorithm}
\usepackage[noend]{algpseudocode}

\usepackage{amssymb}

\title{Single-Stage Heavy-Tailed Food Classification}
\name{Jiangpeng He and Fengqing Zhu}
\address{
Elmore Family School of Electrical and Computer Engineering \\ 
Purdue University, West Lafayette, Indiana, U.S.A.
}
\begin{document}
%
\maketitle
\begin{abstract}
Deep learning based food image classification has enabled more accurate nutrition content analysis for image-based dietary assessment by predicting the types of food in eating occasion images. However, there are two major obstacles to apply food classification in real life applications. First, real life food images are usually heavy-tailed distributed, resulting in severe class-imbalance issue. Second, it is challenging to train a single-stage (\textit{i.e.} end-to-end) framework under heavy-tailed data distribution, which cause the over-predictions towards head classes with rich instances and under-predictions towards tail classes with rare instance. In this work, we address both issues by introducing a novel single-stage heavy-tailed food classification framework. 
Our method is evaluated on two heavy-tailed food benchmark datasets, Food101-LT and VFN-LT, and achieves the best performance compared to existing work with over $5\%$ improvements for top-1 accuracy. 
\end{abstract}


\vspace{-0.2cm}
\section{Introduction}
\label{sec:intro}

\vspace{-0.2cm}
Image-based dietary assessment~\cite{he2020multitask} aims to determine the foods and corresponding nutrition from eating occasion images to enable automated analysis of nutrition intake. Despite significant progress made in food classification by leveraging deep learning models, the performance still struggles when applied in real world applications. One of the major challenges is that heavy-tailed distribution of food classes in real life where a minority of food types are consumed more frequently than the majority of foods, resulting in severe class-imbalance. Therefore, simply training a deep model on static food dataset cannot generalize well in real world. 

\begin{figure}[t]
\begin{center}
  \includegraphics[width=.8\linewidth]{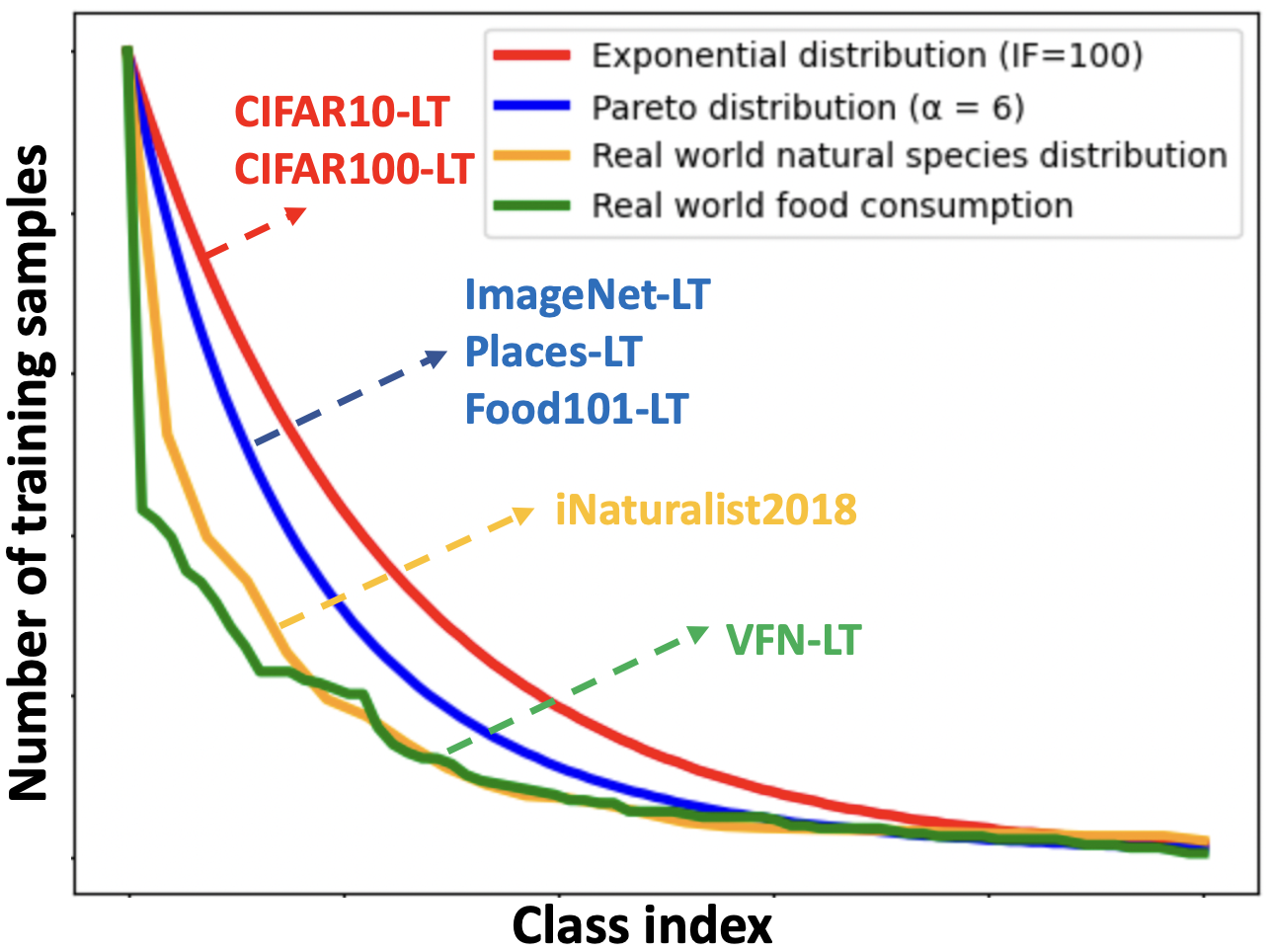} 
  \vspace{-0.4cm}
  \caption{Existing benchmark datasets for long-tailed classification including CIFAR10/100-LT(imbalance factor 100)~\cite{LDAM}, ImageNet-LT~\cite{liu2019large}, Places-LT~\cite{liu2019large}, Food101-LT~\cite{he2022long}, iNaturalist2018~\cite{inaturalist} and VFN-LT~\cite{he2022long}. Note that we normalize the x,y units into the same scale for visualization purpose.}
  \label{fig:intro}
\end{center}
\vspace{-0.7cm}
\end{figure}

Recent work~\cite{he2022long} shows that food consumption in real world follows the heavy-tailed distribution, which contains heavier right skewed tail than exponential distribution~\cite{bryson1974heavy} as shown in Figure~\ref{fig:intro}. One of the major challenges of heavy-tailed classification is the prediction bias towards classes that contain more instances (\textit{i.e. }head classes). From the perspective of learned feature extractor, the data representation of instance-rich classes occupy dominant portion of learned feature space due to the bias of semantic labels~\cite{kang2021exploring}, resulting in less discrimination in feature space (\textit{i.e. }higher inter-class similarity), especially between instance-rich (head) and instance-rare (tail) classes. In addition, from the perspective of learned classifier, the norm of weight vectors in head classes becomes much larger than in tail classes~\cite{weight_balancing}, which outputs imbalanced logits and cause the prediction bias. 

Despite a plethora of studies focused on general heavy-tailed classification tasks, applying these existing strategies to food image classification is not straightforward. Moreover, the complexity of food image classification is amplified due to the increased levels of intra-class dissimilarity and inter-class similarity~\cite{mao2020visual}. In this work, we propose a novel single-stage heavy-tailed food classification framework to address the above mentioned issues. Specifically, we first introduce a new epoch-wise instance sampler to generate balanced training data for each epoch by efficiently under-sampling head classes and over-sampling tail classes with data augmentation. Then, we leverage cosine normalization on the last fully connected layer to obtain scale-invariant output logits. Furthermore, targeting on food images, we construct positive pairs by selecting data from the same class to improve intra-class compactness and construct negative pairs by cross-matching head and tail classes to improve inter-class discrepancy. Our method is evaluated on the Food101-LT and the VFN-LT datasets, both are heavy-tailed distributed and the VFN-LT exhibits real world food consumption pattern. While training in an end-to-end fashion, our method outperforms both existing food recognition and long-tailed classification work with a large margin of over $5\%$ improvements in terms of top-1 classification accuracy. 


\vspace{-0.4cm}
\section{Related Work}
\label{sec: related work}
\vspace{-0.3cm}
\textbf{Food classification.} Much progress have been achieved in the field of food image classification in recent years, with varying scenarios such as fine-grained classification~\cite{large-scale-food} and continual learning~\cite{He_2021_ICCVW, raghavan2023online}, which targets practical applications prominently including image-based dietary assessment. However, few existing works focus on the heavy-tail issue of food class distribution in real world, lacking a generalized solution when training the deep models in severe class-imbalanced data. The most recent work~\cite{he2022long} fill this gap by introducing new benchmarks and a two-stage framework to integrate knowledge distillation and data augmentation. Nevertheless, the performance still struggles and most importantly, such training procedure is not end-to-end so that the accumulated training stages makes it less practical for real world deployment. In this work, we introduce a novel single-stage training framework and significantly improve the performance compared to existing methods. 
\newline
\textbf{Long-tailed classification.} Long-tailed classification has been widely studied over decades~\cite{zhang2021deep}. In this section, we only review existing works that are most relevant to our method. As introduced in Section~\ref{sec:intro}, one of the major issues of long-tailed classification is the prediction bias towards the head classes caused by training with imbalanced semantic labels. \textit{Data re-sampling} based methods aim to create balanced training data. The most common practice is to under-sample the head classes or over-sample the tail classes. However, such naive random over-sampling~\cite{ROS} intensifies the overfitting problem by using repeated training of tail classes and naive random under-sampling\cite{RUS_ROS} causes knowledge loss as part of data from head classes is discarded, resulting in degraded performance. The most recent work~\cite{CMO, he2022long} applies \textit{data augmentation} to mitigate the issues caused by random sampling. In addition, the \textit{loss re-weighting} based method seeks to balance the gradients by assigning proper weights on different classes or training data~\cite{LDAM, FocalLoss, BSLoss, IBLoss}. Nevertheless, these methods improve the tail class accuracy by significantly sacrificing the head class accuracy. The \textit{logit adjustment} based methods directly shift the output logits by using label frequencies~\cite{logit_adjust}, normalizing the classifier's weights~\cite{kang2019decoupling, Liu_2020_CVPR} or applying regularization~\cite{weight_balancing}. In this work, we first address the issue of re-sampling by introducing an efficient epoch-wise instance sampler where the imbalanced degree gradually increases over the training phase. Next, we apply cosine normalization in the last fully connected layer to obtain scale-invariant output and by integrating a new objective loss to further improve the intra-class compactness and inter-class discrepancy. Although cosine normalization has been widely studied~\cite{cosine_normalization, rebalancing}, it has not been applied in heavy-tailed food classification in an end-to-end fashion.


\vspace{-0.3cm}
\section{Method}
\vspace{-0.3cm}
\label{sec:method}
In this section, we illustrate our proposed single-stage heavy-tailed food classification framework including an epoch-wise instance sampler (Section~\ref{subsec: sampler}), a cosine normalization (Section~\ref{subsec: cosine}), which is integrated with a new loss function (Section~\ref{subsec: newloss}) to further address the inter-class similarity and intra-class diversity that that are inherent in food images. 

\vspace{-0.3cm}
\subsection{Epoch-wise Instance Sampler}
\vspace{-0.1cm}
\label{subsec: sampler}
Though instance-balanced sampling is one of the best strategies for learning unbiased feature representation~\cite{zhang2021deep}, existing over-sampling methods intensify the overfitting for tail classes and the under-sampling methods degrade the performance on head classes as described in Section~\ref{sec: related work}. The most recent work~\cite{he2022long} proposed a hybrid under/over-sampling framework depending on whether that class has more/less instances than a fixed threshold value to achieve balanced training data. However, they require an additional pre-training stage to decide which data to retain or discard for each class, making it less practical due to the decoupling of training process. In this work, we address this issue by introducing an efficient epoch-wise instance sampler. Motivated by the recent studies~\cite{frankle2020early, achille2017critical} that earlier training iterations of neural network contributes more towards the final performance, we propose to replace the fixed threshold in~\cite{he2022long} with a dynamic threshold calculated by the sinusoidal Equation~\eqref{eq:sampler}
\begin{equation} \label{eq:sampler}
\vspace{-0.1cm}
\begin{aligned}
\mathcal{T} = N_{max} - \frac{1}{2}(N_{max} - N_{min})(1+cos(\frac{\pi T_i}{T}))
\end{aligned}
\end{equation}
where $N_{max}$ and $N_{min}$ account for the maximum and minimum number of training samples in a heavy-tailed dataset. $T_i$ indicates how many epochs have been performed and $T$ refers to the total epochs. This dynamic threshold $\mathcal{T} \in [N_{min}, N_{max}]$ is monotonically increasing over the training iterations where we perform under/over-sampling to obtain the same number of $\mathcal{T}$ samples per class depending on the $\mathcal{T}$ for each epoch as illustrated in Algorithm~\ref{alg:sampler}. Therefore, the initial smaller threshold $\mathcal{T}$ with smooth increment ensures a more class-balanced data distribution at earlier training stages for establishing the unbiased feature space. Then the rapid increase of $\mathcal{T}$ in the middle of the training stage helps to address the knowledge loss caused by under-sampling. Finally, the neural network is fine-tuned on almost all the training data when $\mathcal{T}$ is close to $N_{max}$. Note that we also integrate data augmentation when performing over-sample on tail classes as in~\cite{CMO}. 


\begin{algorithm}[t]
\caption{Epoch-Wise Instance Sampler}
\begin{flushleft}
    \hspace*{0.02in} {\bf Input:}
    The heavy-tailed datasets $D$ \\
    \hspace*{0.02in} {\bf Input:} 
    The total number of epochs $T$ \\
\end{flushleft}
\vspace{-0.35cm}
\begin{algorithmic}[1]
\State $C \leftarrow \vert D \vert$ \Comment{\small{Total number of classes}}
\For{$c$ = 1, 2, ... $C$ }
\State $I_c \leftarrow \vert c \vert$ \Comment{\small{Number of instance per class}}
\EndFor
\State $N_{min}, N_{max} \leftarrow Min(I), Max(I)$ 
\For{$T_i$ = 1, 2, ... T}
\State $D_{T_i} \leftarrow \emptyset$ \Comment{\small{training data in current epoch}} 
\State $\mathcal{T} \leftarrow$ Equation~\ref{eq:sampler}($T_i$) \Comment{\small{calculate current threshold}} 
\For{$c$ = 1, 2, ... $C$}
\If {$I_c > \mathcal{T} $}  \Comment{\small{random under-sample}} 
\State $D_{T_i} \leftarrow D_{T_i} \cup \textit{Under-sampling}(c, \mathcal{T})$
\Else \Comment{\small{over-sample with augmentation~\cite{CMO}}} 
\State $D_{T_i} \leftarrow D_{T_i} \cup \textit{Over-sampling}(\textit{Aug}(c), \mathcal{T})$
\EndIf
\EndFor
Training epoch $T_i$ with data $D_{T_i}$ begin
\EndFor
\end{algorithmic}
\label{alg:sampler}
\end{algorithm}

\subsection{Cosine Normalization}
\label{subsec: cosine}
\vspace{-0.2cm}
It is common to update deep neural networks using linear classifier and cross-entropy loss, which can be expressed as 
\begin{equation} \label{eq:common}
\begin{aligned}
\mathcal{L}_{ce}(x,y) = - \sum_{i=1}^{C}y_i\times log (\frac{exp(w_i^Tf(x)+b_i)}{\sum_{j}exp(w_j^Tf(x)+b_j)})
\end{aligned}
\end{equation}
where $x$ and $y$ refers to the input image and semantic label with total $C$ classes. $f(\bullet)$ indicates feature extractor, $w_i$ and $b_i$ denote the weight vectors and bias value corresponding to class $i$ in the linear classifier. However, as shown in~\cite{rebalancing, zhang2021deep}, the norm of weight vectors $||w_i||_2$ becomes much larger in head classes with more training data, which contributes the most of gradients to grow the classifier weights during the training process, resulting in the predictions bias in heavy-tailed classification. In this work, we address this issue by applying cosine normalization in the linear classifier as 
\begin{equation} \label{eq:our}
\vspace{-0.1cm}
\begin{aligned}
\mathcal{L}_{ce}(x,y) = - \sum_{i=1}^{C}y_i\times log (\frac{exp(\tau \langle \bar{w_i}, \bar{f}(x)\rangle}{\sum_{j} exp(\tau \langle\bar{w_j}, \bar{f}(x)\rangle)})
\end{aligned}
\vspace{-0.1cm}
\end{equation}
where we remove bias vector $b$ and apply cosine similarity $\langle \bar{v_1}, \bar{v_2}\rangle = v_1^Tv_2$ using $l_2$ normalized weight vectors $\bar{w_i} = \frac{w_i}{||w_i||_2}$ and extracted feature $\bar{f}(x) = \frac{f(x)}{||f(x)||_2}$. The learnable temperature $\tau$ initialized as $1$ is applied to adjust the magnitudes of the loss during training as the value of cosine similarity $\langle \bar{v_1}, \bar{v_2}\rangle$ is constrained to $[-1, 1]$. The cosine normalization project the weights into hyper-sphere space and make prediction by measuring the angle between normalized input and weight vector, which effectively mitigate the scale issue. 
\vspace{-0.15cm}
\subsection{Intra-class Compactness and Inter-class Discrepancy}
\begin{figure}[t]
\begin{center}
  \includegraphics[width=1.\linewidth]{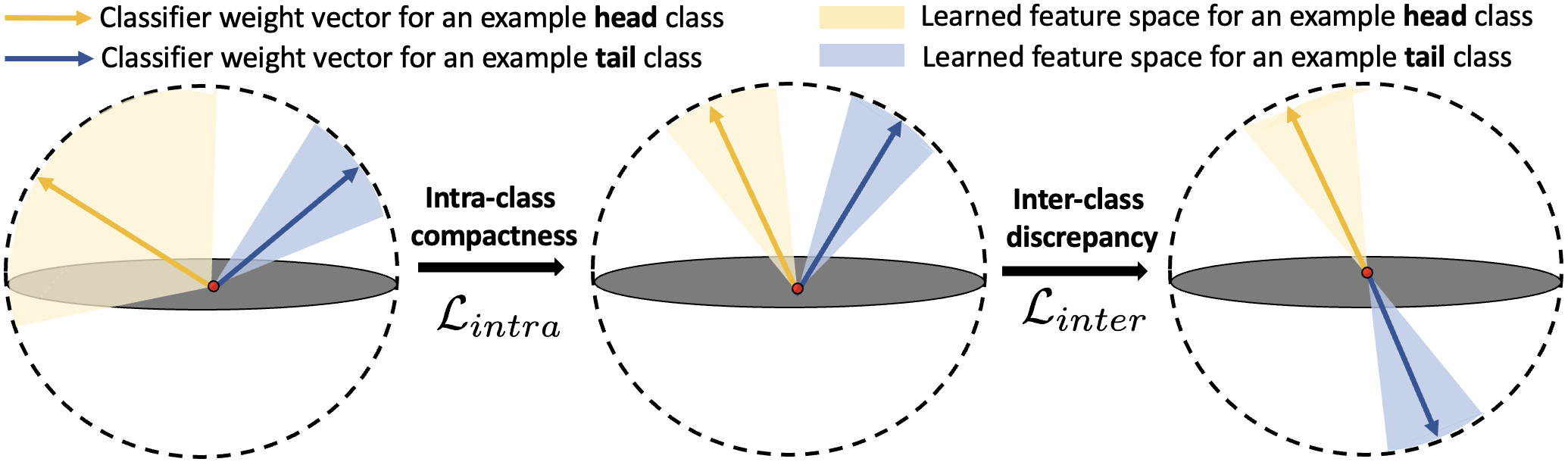}
  \vspace{-0.8cm}
  \caption{Illustration of proposed loss function to improve intra-class compactness and inter-class discrepancy. Note that the cosine normalization project the feature and embeddings into hyper-sphere space.}
  \label{fig:method}
\end{center}
\vspace{-0.5cm}
\end{figure}

\vspace{-0.15cm}
\label{subsec: newloss}
One of the major challenges for food classification is the higher intra-class diversity and inter-class similarity of food images~\cite{mao2020visual}, which becomes more significant in heavy-tailed scenario. Therefore, we propose a novel loss function in this section as illustrated in Figure~\ref{fig:method}, which can be integrated effectively with the cosine normalization described in Section~\ref{subsec: cosine} to improve the intra-class compactness and inter-class discrepancy for food classification. Specifically, given the sampled training data for current epoch $D_{T_i}$ and the entire training set $D$, we first pair each $x \in D_{T_i}$ with an positive image $x_p \in D$ with the same semantic class and maximize the cosine similarity to improve intra-class compactness as
\begin{equation} \label{eq:intra}
\begin{aligned}
\mathcal{L}_{intra}(x) = 1 - \langle \bar{f}(x), \bar{f}(x_p) \rangle
\end{aligned}
\end{equation}
Then, we propose to construct negative pair $(x, x_n)$ by cross matching head and tail classes samples and force $\langle \bar{f}(x), \bar{f}(x_p) \rangle > \langle \bar{f}(x), \bar{f}(x_n) \rangle$ to improve inter-class discrepancy as expressed by
\begin{equation} \label{eq:inter}
\begin{aligned}
\mathcal{L}_{inter}(x) = [\langle \bar{f}(x), \bar{f}(x_n) \rangle - \langle \bar{f}(x), \bar{f}(x_p) \rangle]_{+} 
\end{aligned}
\end{equation}
where $[z]_+ = max(z, 0)$. The final objective loss function for the entire framework is give by 
\begin{equation} \label{eq:final}
\begin{aligned}
\mathcal{L}(x, y) = \mathcal{L}_{ce}(x, y) + \mathcal{L}_{intra}(x) + \mathcal{L}_{inter}(x) 
\end{aligned}
\end{equation}
which can jointly train feature extractor and classifier in single-stage and end-to-end fashion.

\vspace{-0.2cm}
\section{Experiments}
\vspace{-0.2cm}
\label{sec:exp}
In this section, we evaluate our proposed method by comparing with existing work from both food classification and long-tailed classification fields. While focusing on food data, we also show the effectiveness of our method by using general benchmark dataset. Finally, we conduct ablation study to evaluate each component of our proposed method.  

\vspace{-0.3cm}
\subsection{Experimental Setup}
\vspace{-0.1cm}
\label{subsec: expsetup}
\indent \indent  \textbf{Datasets.} We use three benchmarks including Food101-LT~\cite{he2022long}, VFN-LT~\cite{he2022long} and CIFAR100-LT~\cite{LDAM} where the first two are food specific datasets and the last is general task dataset. Following the proposed benchmarks~\cite{he2022long}, Food101-LT has 101 food classes where the number of training data per class vary from $[5, 750]$ with 28 head classes and 73 tail classes, the test set is balanced with 250 images per class. VFN-LT contains 74 food classes with 22 head classes and 52 tail classes, which exhibits real world food consumption with training samples vary from [1, 288] and each class contains 25 test images. CIFAR100-LT is created by applying exponential distribution with different imbalanced factor~\cite{LDAM} (we use 100 in this work) on CIFAR-100~\cite{CIFAR}, which has 100 classes of general objects. 

\textbf{Compared methods.} As the long-tailed classification area evolves rapidly, we compare with the most relevant methods as described in Section~\ref{sec: related work} including \textbf{ROS}~\cite{ROS}, \textbf{RUS}~\cite{RUS_ROS} and \textbf{Food2stage}~\cite{he2022long} for \textit{re-sampling} based,  \textbf{CMO}~\cite{CMO} for \textit{augmentation} based, \textbf{LDAM}~\cite{LDAM}, \textbf{BS}~\cite{BSLoss}, \textbf{IB}~\cite{IBLoss} and \textbf{Focal}~\cite{FocalLoss} loss for \textit{re-weighting} based, \textbf{WB}~\cite{weight_balancing} and \textbf{LA}~\cite{logit_adjust} for \textit{logit adjustment} based methods. Besides we include vanilla training using cross-entropy as \textbf{baseline} and \textbf{HFR}~\cite{mao2020visual} for general food classification task. 

\textbf{Evaluation and implementation details.} We use Top-1 classification accuracy as the evaluation metric and provide the performance on both head and tail classes for results on Food101-LT and VFN-LT. Note that we only provide the overall accuracy on CIFAR100-LT as this work specifically focus on food images. We apply ResNet-18\cite{RESNET} on Food101-LT, VFN-LT and ResNet-32 on CIFAR100-LT. We train 150 epochs with batch size 128 using SGD optimizer, the learning rate starts from $0.01$ and decays with cosine learning rate scheduler. We run each experiment 3 times and report the average performance. 

\begin{table}[t!]
    \centering
    \scalebox{.7}{
    \begin{tabular}{cccccccc}
        \hline
        Datasets & \textbf{CIFAR100-LT} &\multicolumn{3}{c}{\textbf{Food101-LT}} & \multicolumn{3}{c}{\textbf{VFN-LT}} \\
        \hline
        Accuracy(\%) & Overall & Head & Tail & Overall & Head & Tail & Overall \\
        \hline
        Baseline &38.2 &65.8 & 20.9 & 33.4 & 62.3 & 24.4 & 35.8  \\
        HFR~\cite{mao2020visual}&38.7 &\textbf{65.9} & 21.2 & 33.7 & 62.2 & 25.1 & 36.4  \\
        ROS~\cite{ROS} &39.4 & 65.3 & 20.6 & 33.2 & 61.7 & 24.9 & 35.9 \\
        RUS~\cite{RUS_ROS} &37.6 & 57.8 & 23.5 & 33.1 & 54.6 & 26.3 & 34.8 \\
        CMO~\cite{CMO} &43.9 & 64.2 & 31.8 & 40.9 & 60.8 & 33.6 &42.1\\
        LDAM~\cite{LDAM} &43.3  & 63.7 & 29.6 & 39.2 & 60.4 & 29.7 & 38.9 \\
        BS~\cite{BSLoss} &45.6 & 63.9& 32.2 & 41.1 & 61.3 & 32.9 & 41.9 \\
        IB~\cite{IBLoss} &45.2 & 64.1 & 30.2 & 39.7 & 60.2 & 30.8 & 39.6 \\
        Focal~\cite{FocalLoss} &39.2 & 63.9 & 25.8 & 36.5 & 60.1 & 28.3 & 37.8 \\
        Food2stage~\cite{he2022long}  &45.9 & 65.2 & 33.9 & 42.6 & 61.9 & 37.8 & 45.1 \\
        WB~\cite{weight_balancing} &46.3 & 63.8 & 36.2 & 43.9 & 64.5 & 38.8 & 46.4 \\
        LA~\cite{logit_adjust} &43.9 & 60.4 & 37.0 & 43.5 & 60.4 & 39.2 & 45.5 \\
        \hline
        Ours &\textbf{47.6} & 65.7 & \textbf{42.9} & \textbf{49.3} & \textbf{66.0} & \textbf{45.1} & \textbf{51.2} \\     
        \hline
    \end{tabular}
    }
    \vspace{-0.2cm}
    \caption{Top-1 accuracy on CIFAR100-LT(imbalanced factor 100), Food101-LT and VFN-LT. }
    \label{tab:expresult}
    \vspace{-0.3cm}
\end{table}

\vspace{-0.3cm}
\subsection{Results on Benchmark Datasets}
\vspace{-0.2cm}
\label{subsec: expresult}
The experimental results on CIFAR100-LT, Food101-LT and VFN-LT by comparing with existing work are summarized in Table~\ref{tab:expresult}. Overall, our proposed method achieves best performance on both general object dataset and food datasets, and outperforms existing methods with large margins of $5\%$ on both Food101-LT and VFN-LT. We observe heavily biased performance between head and tail classes due to the severe class-imbalance issue as illustrated in the two food image datasets. Though existing methods improve the overall accuracy compared to baseline by either balancing the training data/loss or directly adjusting the classifier's output, the performance on food classification is still limited due to the higher intra-class diversity and inter-class similarity along with heavier-tail than general long-tailed datasets as introduced in Section~\ref{sec:intro}. Our method addresses this by considering the imbalance issue in terms of both learned feature and classifier, which is able to improve the performance without sacrificing the tail classes accuracy.  

\begin{table}[t]
    \centering
    \scalebox{.8}{
    \begin{tabular}{ccc|cccccc}
        \hline
        & &  &\multicolumn{3}{c}{\textbf{Food101-LT}} & \multicolumn{3}{c}{\textbf{VFN-LT}} \\
        EIS & CN & I-Loss& Head & Tail & Overall & Head & Tail & Overall \\
         \hline
         & & & \textbf{65.8} & 20.9 & 33.4 & 62.3 & 24.4 & 35.8 \\
         \checkmark  & & & 64.3 & 40.9 & 47.2 & 62.5 & 40.7 & 47.2 \\
         & \checkmark & & 62.7 & 34.9 & 42.6 & 63.8 & 34.1 & 43.6 \\
         & \checkmark& \checkmark& 65.5 & 39.2 & 46.4 & 65.6 & 39.6 & 47.5 \\
         \checkmark  &\checkmark &\checkmark & 65.7 & \textbf{42.9} & \textbf{49.3} & \textbf{66.0} & \textbf{45.1} & \textbf{51.2} \\
         
        \hline
        \hline
    \end{tabular}
    }
    \vspace{-0.3cm}
    \caption{Ablation study on Food101-LT and VFN-LT. }
    \vspace{-0.2cm}
    \label{tab:ablation}
\end{table}

\begin{figure}[t]
\begin{center}
  \includegraphics[width=1.\linewidth]{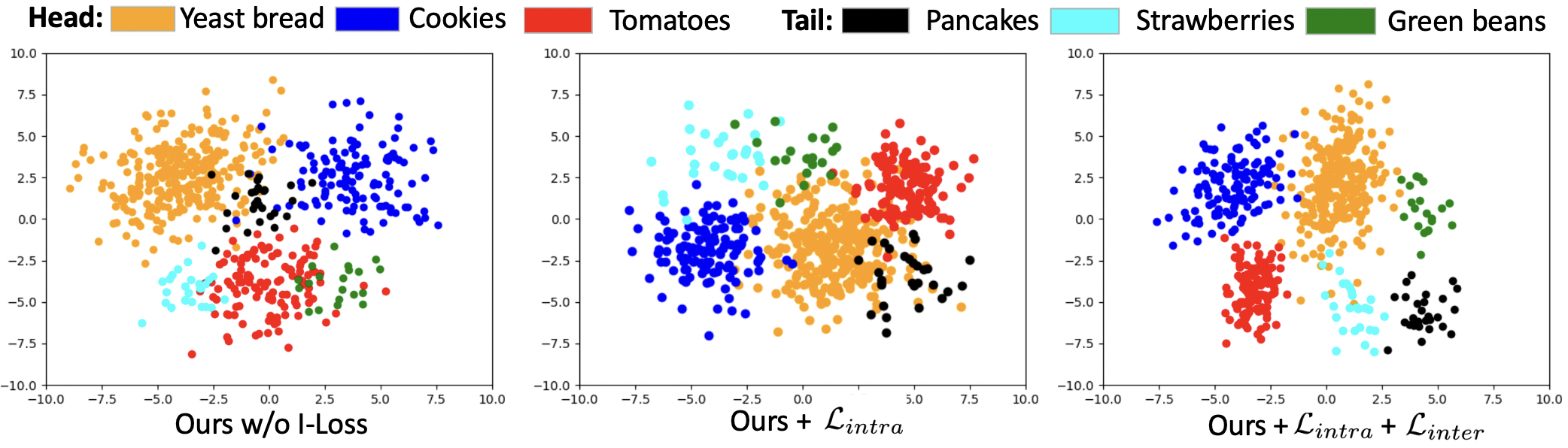}
  \vspace{-0.8cm}
  \caption{The t-SNE visualization on VFN-LT with 3 food classes selected from both head and tail classes, respectively.}
  \label{fig:tsne}
\end{center}
\vspace{-0.6cm}
\end{figure}

\vspace{-0.3cm}
\subsection{Ablation Study}
\vspace{-0.1cm}
\label{subsec: ablation}
In this section, we validate the effectiveness for each component in our framework including: (1) Epoch-wise Instance Sampler (\textbf{EIS}) in Section~\ref{subsec: sampler}, (2) Cosine Normalization (\textbf{CN}) in Section~\ref{subsec: cosine} and (3) the corresponding loss function (\textbf{I-Loss}) in Section~\ref{subsec: newloss}. As shown in Table~\ref{tab:ablation}, the EIS works efficiently to mitigate the class-imbalance issue while losing some knowledge on head classes. In addition, solely applying CN on classifier still result in prediction bias due to the learned imbalanced representation, which is addressed by I-Loss to make features more separable and boost performance. Our method by integrating EIS, CN and I-Loss obtain best accuracy on both head and tail classes. It is also worth noting that the I-Loss also helps to retain and improve the knowledge of head classes as we are using entire training set to construct positive and negative pairs as illustrated in Section~\ref{subsec: newloss}. Finally, we visualize the learned features with selected classes as shown in Fig~\ref{fig:tsne} to validate both $\mathcal{L}_{intra}$ and $\mathcal{L}_{inter}$ in I-Loss. We can readily to see the compactness of features after applying $\mathcal{L}_{intra}$ and become more separable by adding $\mathcal{L}_{inter}$, which further explains our best performance in classifying foods in heavy-tailed distribution. 




\vspace{-0.2cm}
\section{Conclusion and Future Work}
\vspace{-0.2cm}
\label{sec: conclusion}
In this work, we focus on end-to-end food classification in heavy-tailed distribution where a small part of foods are consumed more frequently than others. We first introduce an epoch-wise instance sampler with dynamic threshold which increases over the training iterations to mitigate class-imbalance issue. We then apply cosine normalization on the classifier to obtain scale-invariant output and integrate it with a new loss function to improve the intra-class compactness and inter-class discrepancy. While focusing on food images, our method is evaluated on three benchmark datasets including two food image datasets. Our method achieves the best performance and the ablation study also  validates each component of the proposed method. For future work, we plan to explore heavy-tailed food classification in a more realistic scenario where the data comes sequentially overtime.

\bibliographystyle{IEEEbib}

{\small\bibliography{strings,refs}}

\begin{thebibliography}{10}

\bibitem{he2020multitask}
Jiangpeng He, Zeman Shao, Janine Wright, Deborah Kerr, Carol Boushey, and
  Fengqing Zhu,
\newblock ``Multi-task image-based dietary assessment for food recognition and
  portion size estimation,''
\newblock {\em 2020 IEEE Conference on Multimedia Information Processing and
  Retrieval}, pp. 49--54, 2020.

\bibitem{LDAM}
Kaidi Cao, Colin Wei, Adrien Gaidon, Nikos Arechiga, and Tengyu Ma,
\newblock ``Learning imbalanced datasets with label-distribution-aware margin
  loss,''
\newblock {\em Advances in neural information processing systems}, vol. 32,
  2019.

\bibitem{liu2019large}
Ziwei Liu, Zhongqi Miao, Xiaohang Zhan, Jiayun Wang, Boqing Gong, and Stella
  Yu,
\newblock ``Large-scale long-tailed recognition in an open world,''
\newblock {\em Proceedings of IEEE conference on computer vision and pattern
  recognition}, pp. 2537--2546, 2019.

\bibitem{he2022long}
Jiangpeng He, Luotao Lin, Heather~A Eicher-Miller, and Fengqing Zhu,
\newblock ``Long-tailed food classification,''
\newblock {\em Nutrients}, vol. 15, no. 12, pp. 2751, 2023.

\bibitem{inaturalist}
Grant Van~Horn, Oisin Mac~Aodha, Yang Song, Yin Cui, Chen Sun, Alex Shepard,
  Hartwig Adam, Pietro Perona, and Serge Belongie,
\newblock ``The inaturalist species classification and detection dataset,''
\newblock {\em Proceedings of the IEEE conference on computer vision and
  pattern recognition}, pp. 8769--8778, 2018.

\bibitem{bryson1974heavy}
Maurice~C Bryson,
\newblock ``Heavy-tailed distributions: properties and tests,''
\newblock {\em Technometrics}, vol. 16, no. 1, pp. 61--68, 1974.

\bibitem{kang2021exploring}
Bingyi Kang, Yu~Li, Sa~Xie, Zehuan Yuan, and Jiashi Feng,
\newblock ``Exploring balanced feature spaces for representation learning,''
\newblock {\em International Conference on Learning Representations}, 2021.

\bibitem{weight_balancing}
Shaden Alshammari, Yu-Xiong Wang, Deva Ramanan, and Shu Kong,
\newblock ``Long-tailed recognition via weight balancing,''
\newblock {\em Proceedings of the IEEE/CVF Conference on Computer Vision and
  Pattern Recognition}, pp. 6897--6907, 2022.

\bibitem{mao2020visual}
Runyu Mao, Jiangpeng He, Zeman Shao, Sri~Kalyan Yarlagadda, and Fengqing Zhu,
\newblock ``Visual aware hierarchy based food recognition,''
\newblock {\em Proceedings of the International Conference on Pattern
  Recognition Workshop}, pp. 571--598, February 2021.

\bibitem{large-scale-food}
Weiqing Min, Zhiling Wang, Yuxin Liu, Mengjiang Luo, Liping Kang, Xiaoming Wei,
  Xiaolin Wei, and Shuqiang Jiang,
\newblock ``Large scale visual food recognition,''
\newblock {\em IEEE Transactions on Pattern Analysis and Machine Intelligence},
  pp. 1--18, 2023.

\bibitem{He_2021_ICCVW}
Jiangpeng He and Fengqing Zhu,
\newblock ``Online continual learning for visual food classification,''
\newblock {\em Proceedings of the IEEE/CVF International Conference on Computer
  Vision Workshops}, pp. 2337--2346, October 2021.

\bibitem{raghavan2023online}
Siddeshwar Raghavan, Jiangpeng He, and Fengqing Zhu,
\newblock ``Online class-incremental learning for real-world food
  classification,''
\newblock {\em arXiv preprint arXiv:2301.05246}, 2023.

\bibitem{zhang2021deep}
Yifan Zhang, Bingyi Kang, Bryan Hooi, Shuicheng Yan, and Jiashi Feng,
\newblock ``Deep long-tailed learning: A survey,''
\newblock {\em arXiv preprint arXiv:2110.04596}, 2021.

\bibitem{ROS}
Jason Van~Hulse, Taghi~M Khoshgoftaar, and Amri Napolitano,
\newblock ``Experimental perspectives on learning from imbalanced data,''
\newblock {\em Proceedings of the 24th international conference on Machine
  learning}, pp. 935--942, 2007.

\bibitem{RUS_ROS}
Mateusz Buda, Atsuto Maki, and Maciej~A Mazurowski,
\newblock ``A systematic study of the class imbalance problem in convolutional
  neural networks,''
\newblock {\em Neural networks}, pp. 249--259, 2018.

\bibitem{CMO}
Seulki Park, Youngkyu Hong, Byeongho Heo, Sangdoo Yun, and Jin~Young Choi,
\newblock ``The majority can help the minority: Context-rich minority
  oversampling for long-tailed classification,''
\newblock {\em Proceedings of the IEEE/CVF Conference on Computer Vision and
  Pattern Recognition}, pp. 6887--6896, 2022.

\bibitem{FocalLoss}
Tsung-Yi Lin, Priya Goyal, Ross Girshick, Kaiming He, and Piotr Doll{\'a}r,
\newblock ``Focal loss for dense object detection,''
\newblock {\em Proceedings of the IEEE international conference on computer
  vision}, pp. 2980--2988, 2017.

\bibitem{BSLoss}
Jiawei Ren, Cunjun Yu, Xiao Ma, Haiyu Zhao, Shuai Yi, et~al.,
\newblock ``Balanced meta-softmax for long-tailed visual recognition,''
\newblock {\em Advances in neural information processing systems}, vol. 33, pp.
  4175--4186, 2020.

\bibitem{IBLoss}
Seulki Park, Jongin Lim, Younghan Jeon, and Jin~Young Choi,
\newblock ``Influence-balanced loss for imbalanced visual classification,''
\newblock {\em Proceedings of the IEEE/CVF International Conference on Computer
  Vision}, pp. 735--744, 2021.

\bibitem{logit_adjust}
Aditya~Krishna Menon, Sadeep Jayasumana, Ankit~Singh Rawat, Himanshu Jain,
  Andreas Veit, and Sanjiv Kumar,
\newblock ``Long-tail learning via logit adjustment,''
\newblock {\em International Conference on Learning Representations}, 2021.

\bibitem{kang2019decoupling}
Bingyi Kang, Saining Xie, Marcus Rohrbach, Zhicheng Yan, Albert Gordo, Jiashi
  Feng, and Yannis Kalantidis,
\newblock ``Decoupling representation and classifier for long-tailed
  recognition,''
\newblock {\em International Conference on Learning Representations}, 2020.

\bibitem{Liu_2020_CVPR}
Jialun Liu, Yifan Sun, Chuchu Han, Zhaopeng Dou, and Wenhui Li,
\newblock ``Deep representation learning on long-tailed data: A learnable
  embedding augmentation perspective,''
\newblock {\em Proceedings of the IEEE Conference on Computer Vision and
  Pattern Recognition.}, June 2020.

\bibitem{cosine_normalization}
Chunjie Luo, Jianfeng Zhan, Xiaohe Xue, Lei Wang, Rui Ren, and Qiang Yang,
\newblock ``Cosine normalization: Using cosine similarity instead of dot
  product in neural networks,''
\newblock {\em Artificial Neural Networks and Machine Learning}, pp. 382--391,
  2018.

\bibitem{rebalancing}
Saihui Hou, Xinyu Pan, Chen~Change Loy, Zilei Wang, and Dahua Lin,
\newblock ``Learning a unified classifier incrementally via rebalancing,''
\newblock {\em Proceedings of the IEEE Conference on Computer Vision and
  Pattern Recognition}, pp. 831--839, 2019.

\bibitem{frankle2020early}
Jonathan Frankle, David~J Schwab, and Ari~S Morcos,
\newblock ``The early phase of neural network training,''
\newblock {\em arXiv preprint arXiv:2002.10365}, 2020.

\bibitem{achille2017critical}
Alessandro Achille, Matteo Rovere, and Stefano Soatto,
\newblock ``Critical learning periods in deep neural networks,''
\newblock {\em arXiv preprint arXiv:1711.08856}, 2017.

\bibitem{CIFAR}
Alex Krizhevsky, Geoffrey Hinton, et~al.,
\newblock ``Learning multiple layers of features from tiny images,''
\newblock {\em Technical Report}, 2009.

\bibitem{RESNET}
Kaiming He, Xiangyu Zhang, Shaoqing Ren, and Jian Sun,
\newblock ``Deep residual learning for image recognition,''
\newblock {\em Proceedings of the IEEE Conference on Computer Vision and
  Pattern Recognition}, pp. 770--778, 2016.

\end{thebibliography}

\end{document}